\def\year{2022}\relax
\documentclass[letterpaper]{article} 
\usepackage{aaai22}  
\usepackage{times}  
\usepackage{helvet}  
\usepackage{courier}  
\usepackage[hyphens]{url}  
\usepackage{graphicx} 
\usepackage{natbib}  
\usepackage{caption} 
\DeclareCaptionStyle{ruled}{labelfont=normalfont,labelsep=colon,strut=off} 
\usepackage{algorithm}
\usepackage{algorithmic}
\usepackage{subfigure}
\usepackage{bbm}
\usepackage{amsmath}
\usepackage{amssymb}
\usepackage{comment}
\newcommand*\samethanks[1][\value{footnote}]{\footnotemark[#1]}

\usepackage{newfloat}
\usepackage{listings}
\floatstyle{ruled}
\newfloat{listing}{tb}{lst}{}
\floatname{listing}{Listing}
\title{Highlighting Object Category Immunity for the Generalization of Human-Object Interaction Detection}
\author {
    Xinpeng Liu\thanks{The first two authors contribute equally.},
    Yong-Lu Li\samethanks,
    Cewu Lu\thanks{Cewu Lu is corresponding author, member of Qing Yuan Research Institute and MoE Key Lab of Artificial Intelligence, AI Institute, Shanghai Jiao Tong University, China and Shanghai Qi Zhi institute.}
}
\affiliations {
    Shanghai Jiao Tong University\\
    xinpengliu0907@gmail.com, \{yonglu\_li, lucewu\}@sjtu.edu.cn
}

\begin{document}

\maketitle

\begin{abstract}
Human-Object Interaction (HOI) detection plays a core role in activity understanding. 
As a compositional learning problem (human-verb-object), studying its generalization matters.
However, widely-used metric mean average precision (mAP) maybe not enough to model the compositional generalization well.
Here, we propose a novel metric, \textbf{mPD (mean Performance Degradation)}, as a complementary of mAP to evaluate the performance gap among compositions of \textit{different objects} and the \textit{same verb}. 
Surprisingly, mPD reveals that previous state-of-the-arts usually do not generalize well.
With mPD as a cue, we propose \textbf{Object Category (OC) Immunity} to advance HOI generalization. 
Concretely, our core idea is to prevent model from learning \textit{spurious object-verb correlations} as a short-cut to over-fit the train set.
To achieve OC-immunity, we propose an OC-immune network that decouples the inputs from OC, extracts OC-immune representations and leverages uncertainty quantification to generalize to unseen objects.
In both conventional and zero-shot experiments, our method achieves decent improvements. 
To fully evaluate the generalization, we design a new and more difficult benchmark, on which we present significant advantage.
The code is available at \textit{https://github.com/Foruck/OC-Immunity}.

\end{abstract}

\section{Introduction}
\label{intro}
Human-Object Interaction (HOI) detection recently attracts enormous attention. 
It is generally defined as detecting $\langle human,verb,object \rangle$ triplets~\cite{hicodet} from still images, which is a sub-task of visual relationship detection~\cite{Lu2016Visual,partstate}. 
It plays an important role in robot manipulation~\cite{hayes2017interpretable}, surveillance event detection~\cite{abnormal,unusualeventdetection}, trajectory prediction~\cite{sun2021three,sun2020recursive}, video understanding~\cite{pang2020complex,pang2021pgt}, etc.

\begin{figure}
    \centering
    \includegraphics[width=0.45\textwidth]{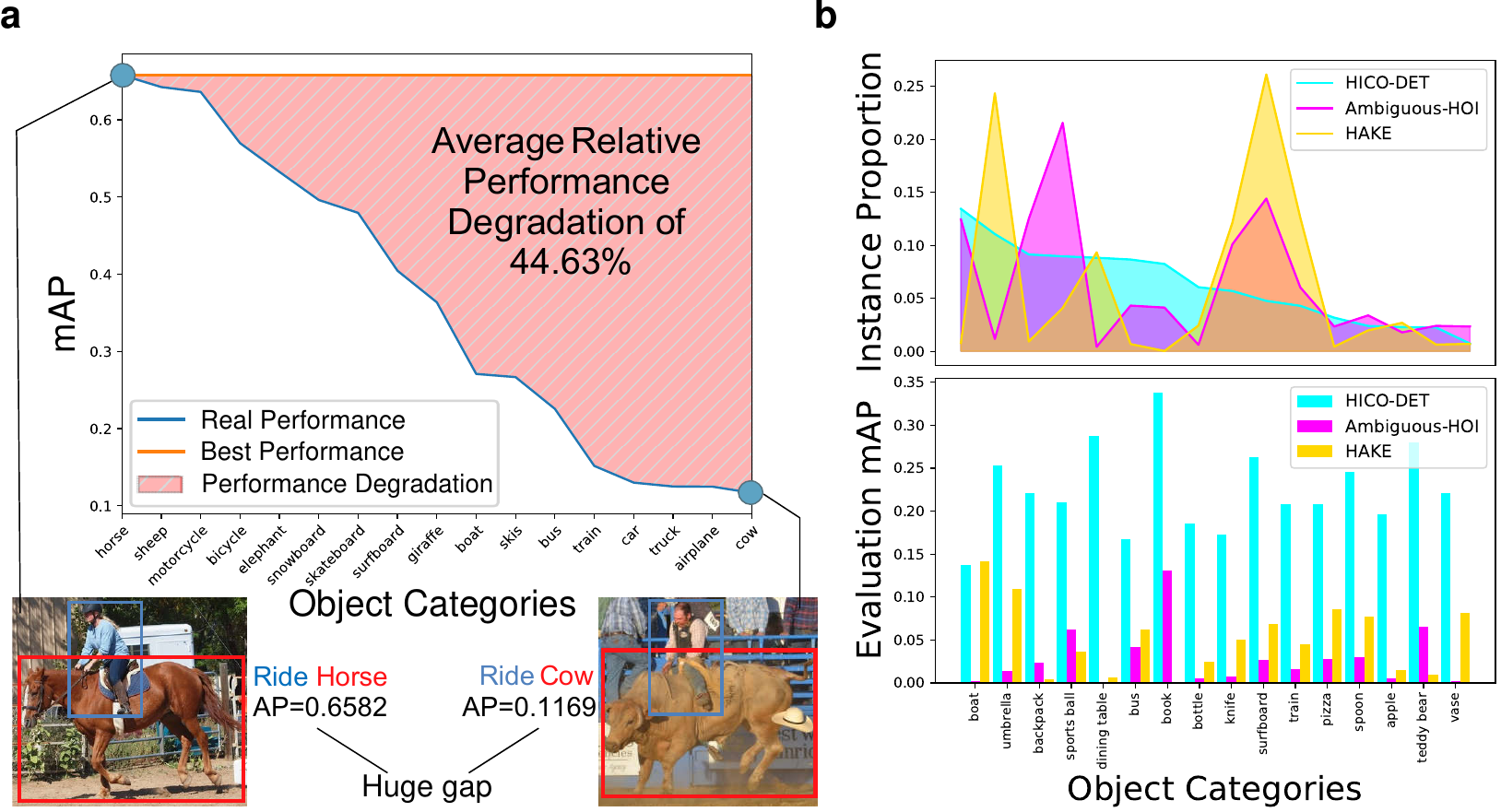}
    \caption{Previous HOI learning methods usually generalize poorly (a) on different object categories; (b) on test set with different object category (OC) distributions. (a). mPD visualization of VCL~\cite{vcl} on verb \textit{ride}. Lower mPD indicates better generalization. (b) Object category distribution and corresponding VCL mAP of different datasets \citet{hicodet,djrn,li2019hake}.}
    \label{Figure:insight}
  \vspace{-0.3cm}
\end{figure}

Recently, impressive progress has been made in this field~\cite{interactiveness,vcl,analogy,djrn}, and the most widely-used metric mAP (mean Average Precision) has reached an impressive level. However, the compositional generalization problem is still open: they provide limited performance on test samples with the \textbf{same} verbs but \textbf{rarely seen} or \textbf{unseen} object categories. 
To illustrate this phenomenon more clearly, we propose a novel metric to directly measure generalization in HOI learning, named \textbf{mPD (mean Performance Degradation)}. In detail, for a given verb and its available objects, we compute the \textit{relative performance gap} between the best and the rest verb-object compositions (i.e., the lower, the better). 
As illustrated in Fig.~\ref{Figure:insight}a, previous methods~\cite{vcl} usually fail to achieve satisfying mPD, resulting in limited generalization: polarized performances on datasets with diverse object distributions, as shown in Fig.~\ref{Figure:insight}b. 

Previous methods usually seek HOI generalization from compositional learning~\cite{vcl,functional}. 
A common way is to import novel objects via language priors~\cite{analogy,wang2020discovering,functional} and model the similarity between seen and unseen categories, then translate $\langle verb$-$(seen),$ $object$-$(seen) \rangle$ to $\langle verb$-$(seen),$ $object$-$(unseen) \rangle$. 
This is intuitive but maybe not enough to process the enormous variety: object categories are inexhaustible. 
Instead, we propose a new perspective: \textbf{object category (OC) immunity}: the performance gap among compositions of \textit{different} objects for the \textit{same} verb is a kind of OC-related \textit{bias}. 
Due to the imbalanced verb-object distribution, models could learn spurious verb-object correlation as a shortcut to over-fit the training set.
To avoid this, we prevent our model from relying too heavily on OC information. 
This enables it to generalize better to rare/unseen objects. 
In other words, we adopt OC-immunity to \textbf{trade-off} between fitting and generalization.

In light of this, we propose a new HOI learning paradigm to improve generalization.
\textbf{(1)} We introduce OC-immunity, with which the model could rely less on object category and provide better performance for unfamiliar objects. 
In detail, firstly, we \textbf{disentangle} the \textit{inputs} of the multi-stream structure~\cite{hicodet,interactiveness} from object category.
Then, each stream would perform verb classification and \textbf{uncertainty quantification}~\cite{kendall2017uncertainties} concurrently, enabling them to \textit{avoid overconfidence when they do not know}. 
Thus, they are expected to be less confident about their mistakes, which is meaningful when encountering unseen objects.
\textbf{(2)} For \textit{object feature} inherently holding category information, we design an OC-immune method via synthesizing object features of different categories as one to mix the category information, thus mitigating the spurious verb-object correlations.
Meanwhile, given the compositional characteristic of HOI, the prediction combination of multi-stream is crucial. So we propose \textbf{calibration-aware unified inference} to exploit the unique advantages of different streams. 
That is, first calibrating streams separately by a variant of Platt scaling~\cite{platt1999probabilistic} and then combining multi-prediction regarding uncertainty. 
\textbf{(3)} As an extra benefit, uncertainty quantification provides an option to utilize data \textbf{without HOI labels}, which can reduce bias and advance generalization. 
For evaluation, we conduct experiments on HICO-DET~\cite{hicodet} under both conventional~\cite{hicodet} and zero-shot settings~\cite{Shen2018Scaling, vcl}, showing considerable improvements. 
To further demonstrate the efficacy of our method, we design a benchmark for HOI generalization, on which impressive advantage is also achieved.

Our contribution includes: 
1) We propose mean performance degradation (mPD) to quantify HOI generalization. 
2) Object category immunity is introduced to HOI learning with a novel paradigm. 
3) A novel benchmark is devised to facilitate researches on HOI generalization. 
4) Our proposed method achieves impressive improvements for both conventional and zero-shot HOI detection.
\section{Related Works}

\noindent{\bf HOI Learning:} 
Large datasets~\cite{hicodet,vcoco,OpenImages} have been released.
Meanwhile, many deep learning-based methods~\cite{Gkioxari2017Detecting,djrn,analogy,vcl,pastanet,fang2021decaug,fang2021dirv,fang2018pairwise} have been proposed.
\citet{hicodet} proposed multi-stream framework followed by subsequent works~\cite{gao2018ican,interactiveness,DRG,vcl}.
\citet{gpnn} and \citet{wang2020contextual} used graphical model to address HOI detection.
\citet{Gkioxari2017Detecting} estimated the interacted object locations.
\citet{gao2018ican} and \citet{pmfnet} adopted self-attention to correlate the human, object, and context.
\citet{interactiveness} modeled interactiveness to suppress non-interactive pairs given human pose~\cite{fang2017rmpe,li2019crowdpose}. 
\citet{djrn} used 3D human information~\cite{you2021understanding} to enhance HOI learning, while \citet{vcl} exploited the compositional characteristic of HOI. 
Also, some works~\cite{analogy,kim2020detecting,zhong2020polysemy} utilized the relationship between HOIs.
\citet{ppdm} and \citet{uniondet} directly detected HOI pairs.

\noindent{\bf Zero-shot HOI Detection} has become a new focus recently~\cite{Shen2018Scaling,functional,analogy,vcl,wang2020discovering}. Shen~\cite{Shen2018Scaling} first proposed to factorize HOI into verb classification and object classification. Some works~\cite{functional,analogy,wang2020discovering} utilized language prior to reason about the relationship between objects for zero-shot generalization. \citet{vcl} made use of the compositional characteristic of HOI to generate novel HOI types. As described, most of them adopt object knowledge to reason about objects, suffering from inexhaustibility.

\noindent{\bf Uncertainty Quantification} models what a model does not know. There has been increasing literature in uncertainty estimation of deep learning models~\cite{bishop1995neural,kendall2017uncertainties,gal2016dropout,lee2020gradients,blundell2015weight,lee2015m}. Some methods~\cite{bishop1995neural,blundell2015weight} are based on Bayesian Neural Network, estimating the distribution of network parameters and producing Bayesian approximation of epistemic uncertainty. MC-Dropout~\cite{gal2016dropout} sampled a discrete model from Bayesian parameter distribution. Moreover, \citet{kendall2017uncertainties} used direct estimation for aleatoric uncertainty and MC-Dropout for epistemic uncertainty. There has also been ensemble-based methods~\cite{lee2015m} and gradient-based methods~\cite{lee2020gradients}. In this paper, we adopt the aleatoric uncertainty estimation in \cite{kendall2017uncertainties}.

\noindent{\bf Calibration} aims to improve the quality of the output confidence of a model. Platt scaling~\cite{platt1999probabilistic} has been shown effective while simple. In this work, we adopt Platt scaling to calibrate the output probability of our model.
\section{Method}
\label{sec:method}
\begin{figure*}[!ht]
	\begin{center}
		\includegraphics[width=0.8\textwidth]{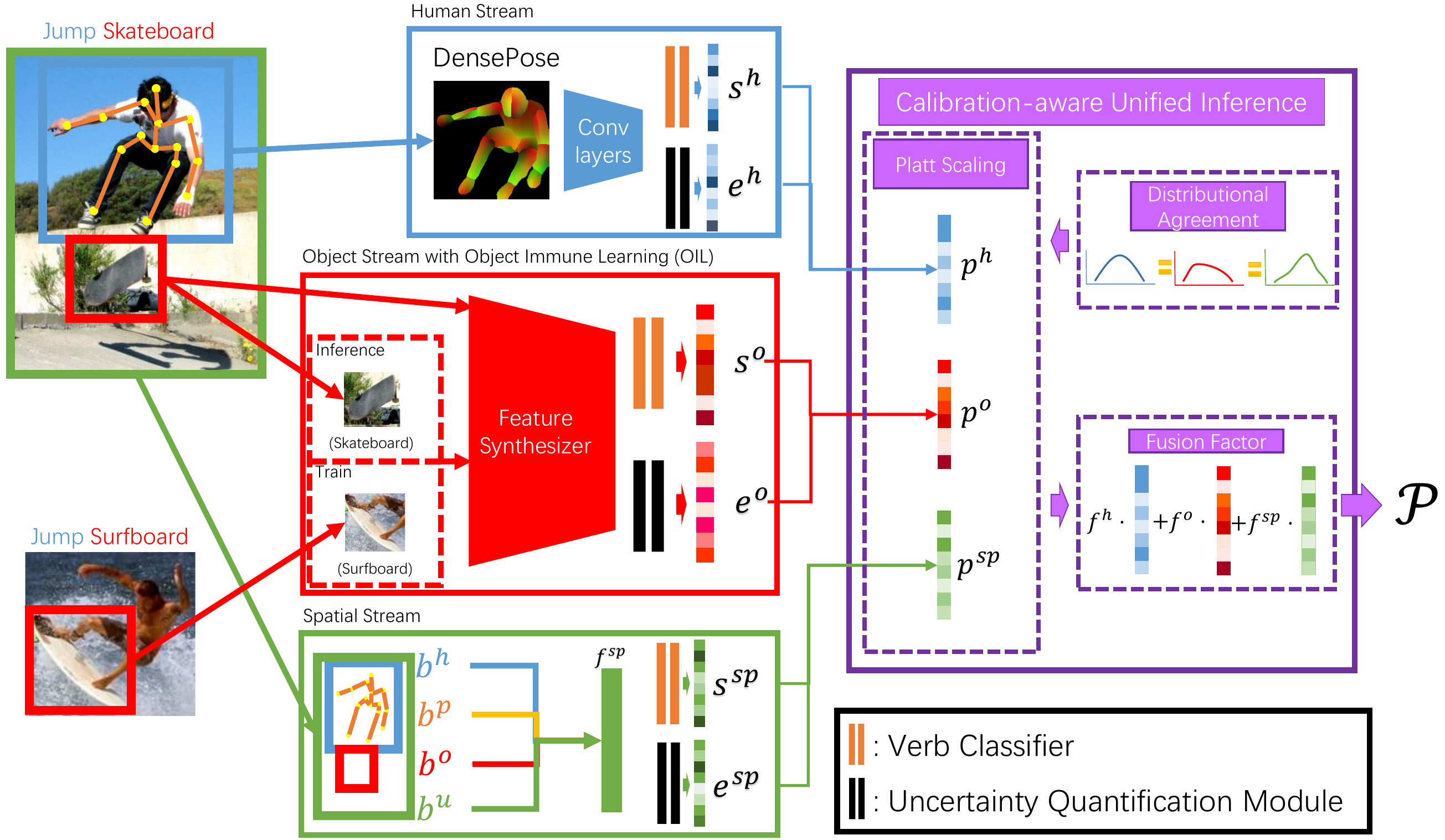}
	\end{center}
	\caption{Overview of our pipeline. We design a multi-stream structure facilitated with uncertainty quantification, then calibration-aware unified inference unifies multi-stream into the final prediction $\mathcal{P}$. $s^h,s^o,s^{sp}$ are the verb classification logits, and $e^h, e^o, e^{sp}$ are the predicted log variances.
	}
	\label{Figure:overview}
	\vspace{-0.4cm}
\end{figure*}

In this section, we first formulate the proposed HOI generalization metric. 
With the metric, our goal is improving generalization by introducing OC-immunity to HOI learning. 
To this end, we modify the multi-stream structure~\cite{interactiveness,hicodet} to realize OC-immunity by disentangling the inputs and devising a new OC-immune learning method.
Furthermore, we utilize uncertainty quantification to unify multi-stream concerning calibration.
\vspace{-0.3cm}
\subsection{Generalization Quantification in HOI}
\label{sec:fair-metric}
Generalization has been a central obstacle to HOI understanding. 
Though previous methods~\cite{djrn,vcl} provide impressive results of mAP, the generalization performance is still limited in transfer experiments and zero-shot settings~\cite{vcl,functional,analogy,Shen2018Scaling}. 
In other words, the relation between mAP and the generalization ability is not as explicit as we expect. 
This reminds us that widely-used mAP might not be sufficient to evaluate the performance of HOI detection, especially for generalization.
To quantify the generalization ability of an HOI learning model, we propose a novel metric, mPD (\textbf{m}ean \textbf{P}recision \textbf{D}egradation). 
For verb $v\in V$, we denote the object categories available for $v$ as $O_v$, $AP_{\langle v,o\rangle}$ as the average precision for HOI composition $\langle v, o\rangle$. Then mPD is formulated as:
\begin{eqnarray}
   \label{eq:mPD}
    mPD &=& \frac{1}{|V|}\mathop{\sum}_{v\in V} \frac{AP_{\langle v, o_{max}\rangle}-\bar{AP_v}}{AP_{\langle v, o_{max}\rangle}}, 
\end{eqnarray}
where $o_{max}$=$\mathop{\textrm{argmax}}_o AP_{\langle v, o\rangle}$,  $\bar{AP_v}$=$\frac{1}{|O_v|}\mathop{\sum}_{o\in O_v}AP_{\langle v, o\rangle}$ is the mean AP for compositions $\langle v, o\rangle$ with $o \in O_v$. mPD measures performance gap between the best-detected composition and the rest. The higher the mPD is, the larger performance gap a model might present on different objects, in other words, worse generalization. 
As shown in Fig.~\ref{Figure:insight}a and Fig.~\ref{Figure:mpd}, mPD reveals previous method~\cite{vcl} has huge performance gap among \textit{different} objects with the \textit{same} verb, which limits generalization.
To address this, we utilize OC-immunity as a proxy to narrow the generalization gap.
\subsection{OC-Immune Network}
\label{sec:protected-attribute}
The performance gap could result from spurious object-verb correlations learned by model as a short-cut to over-fit the training set.  
Previous methods intend to achieve generalization by taking more object categories (OC) into consideration with the help of language priors~\cite{vcl,analogy,functional}, learn the unseen objects via their similarity with the seen objects. 
However, there exists an inherent limitation: We can not exhaust all the objects due to their variety and visual diversity. 
While we take this from another view: an \textbf{inherently OC-immune representation} of HOI. 
The idea is to encourage the model to focus more on interaction-related information instead of OC-related information, thus \textit{weakening} the influence of object categories.
To achieve this, we resort to the multi-stream structure in HOI learning~\cite{hicodet,interactiveness}. 
Despite the relatively poor generalization of previous methods, we find this structure has an inherent relation to OC-immunity. 

In the multi-stream structure, each stream is designed to perform inference based on one specific element: human, object, or spatial configuration, while the human stream and spatial stream are inherently object-immune by \textit{ideal} design.
However, in previous works, this structure performs poorly as an OC-immune model (Fig.~\ref{Figure:insight}a). 
This is because though the separated streams seem to ignore the object category information, the features used are still \textbf{entangled with object}. 
Previous works~\cite{hicodet,gao2018ican,interactiveness} usually adopt pre-trained object detectors~\cite{faster} as feature extractor, and utilize ROI pooling to get features for different elements. 
Some of them~\cite{gao2018ican,interactiveness} also enhance spatial stream with human ROI feature. 
These violate the object-immune principle in two aspects. 
First, the object detector extracted features are trained to classify object categories, e.g., COCO~\cite{coco} 80 objects, thus inevitably strongly correlated to OC. 
Second, human and object boxes in HOI pairs usually overlap, therefore, ROI pooling may introduce OC information to human features. 
To address these violations, we disentangle the feature used by each stream and design different structures for each stream correspondingly, as shown in Fig.~\ref{Figure:overview}. 
Especially, since the object feature inherently holds category information, a new OC-immune learning method is designed for object stream.

\begin{figure}
	\centering
	\subfigure[Human feature involving HOI $\it{ride\ bicycle}$, extracted by ROI Pooling and DensePose. The more a human instance overlaps with an object, the brighter it is painted.]{
		\begin{minipage}[b]{0.43\textwidth}
			\includegraphics[width=\textwidth]{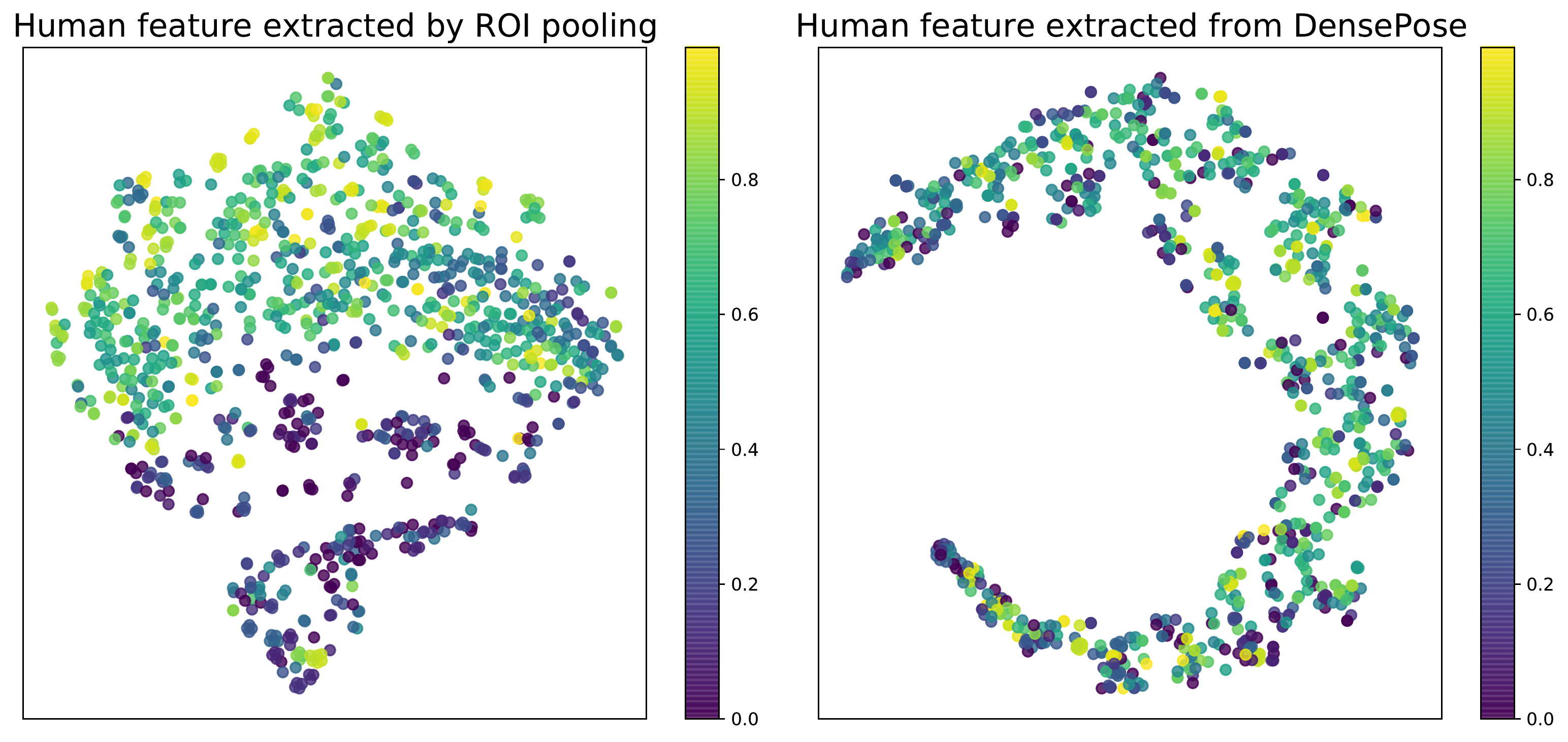}
		\end{minipage}
		\label{fig:feat-hum}
	}
	\subfigure[Object feature of $\it{banana},\ \it{knife},\ \it{tennis\ racket}$, from COCO pre-trained Faster-RCNN and the last FC of our verb classifier. Different colors indicate different object categories.]{
		\begin{minipage}[b]{0.43\textwidth} 
            \includegraphics[width=\textwidth]{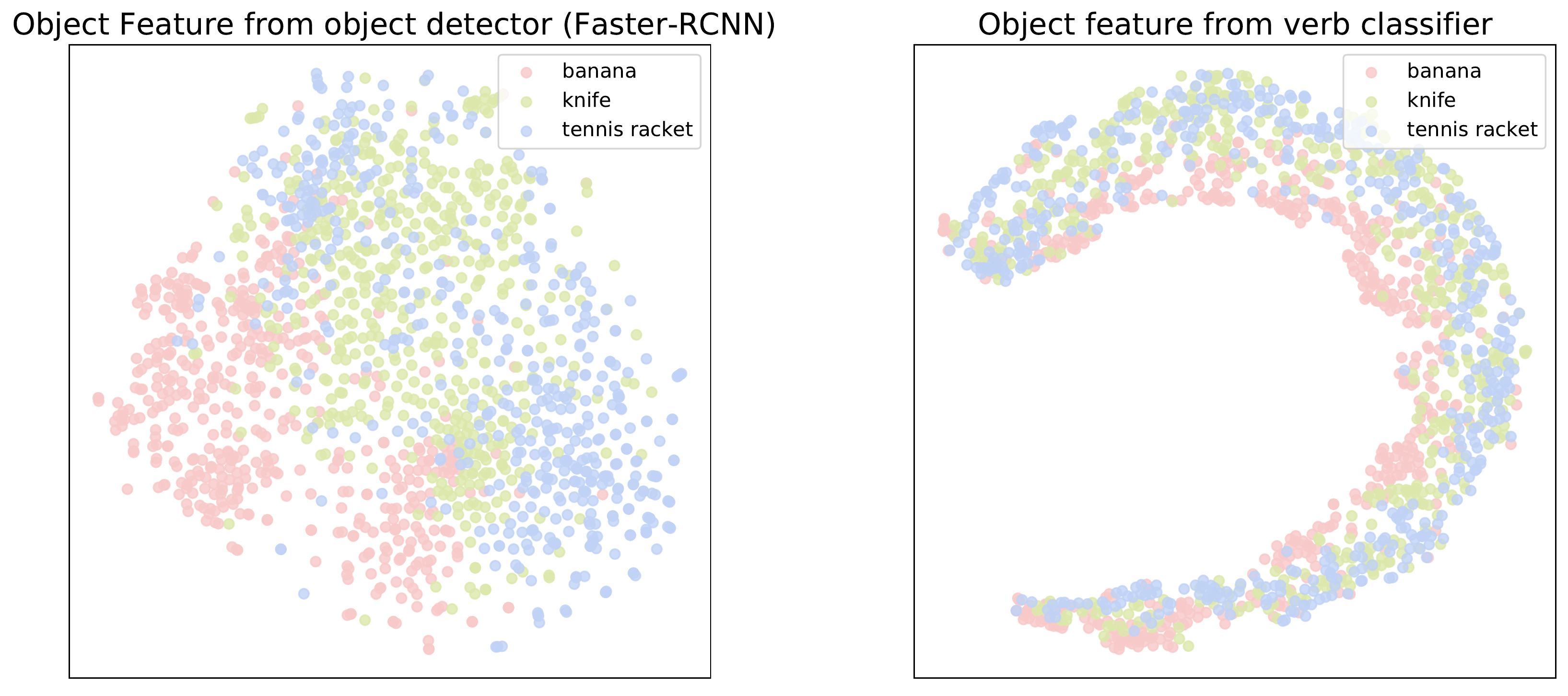}
		\end{minipage}
		\label{fig:feat-obj}
	}
    \vspace{-0.3cm}
	\caption{t-SNE~\cite{tsne} visualization of differently extracted features.} 
\vspace{-0.35cm}
\end{figure}

\noindent{\bf Human Stream} 
For human stream, we use COCO~\cite{coco} pre-trained DensePose~\cite{alp2018densepose} to extract human feature. It provides a pure geometric encoding of human information, which is totally \textbf{object-immune}. 
As shown in Fig.~\ref{fig:feat-hum}, the conventional ROI pooling feature and DensePose feature distribute diversely, while DensePose feature is more robust to the overlap with objects. 
We use multiple convolution layers with batch normalization and ReLU activation, followed by global average pooling to encode the input into features, then employ an MLP to perform verb classification. 

\noindent{\bf Object Stream}
To obtain OC-immune representations, we design a new OC-immune learning method including an object feature synthesizer, an object classifier, and a verb classifier, which are all MLP-based.
With two object ROI features as input, the synthesizer outputs a synthesized feature, which is expected to be the \textbf{intermediate} of the two inputs. 
That is, for example, if we employ the synthesizer to synthesize an object feature with object category labels of ``apple'' and another with ``pear'', the expected result that object classifiers provide for this synthesized vector should be \textbf{0.5} for ``apple'' and ``pear''. 
To achieve this, we first train an object classifier with original data. 
Then, we freeze the object classifier and train the synthesizer to follow the above rules.
Finally, we train our verb classifier with the synthesizer. 
In detail, for a training object feature, we would randomly \textit{duplicate} it or sample another object feature of \textit{similar} object categories (those could be imposed \textit{similar} verbs, e.g., eating apple and banana), then feed them to the synthesizer and verb classifier, use their \textbf{intermediate} verb label as supervision.
In inference, the object feature would be \textbf{duplicated} and fed into the synthesizer and the verb classifier. 
The idea is to expose the verb classifier to feature with \textit{ambiguous} or \textit{corrupted} object category information, while still keeping the verb semantics within the feature. 
This weakens the correlation between verb and OC that could be perceived by the verb classifier, encouraging the classifier to resort to other clues. 
Thus, the influence imposed by OC on the verb classifier would be weakened.
To show the immunity of our \textit{verb} classifier, we visualize features extracted by COCO~\cite{coco} pre-trained Faster-RCNN~\cite{faster} and the last FC of our verb classifier in Fig.~\ref{fig:feat-obj}. 
As illustrated, the latter is less correlated with OC than the former. 

\noindent{\bf Spatial Stream} 
Previously, some works~\cite{hicodet,interactiveness} directly concatenate human features with spatial configuration features for spatial stream, which violates the object-immune principle as stated before.
Therefore, we employ only human and object bounding boxes $b^h, b^o \in \mathbb{R}^{2\times2}$ and 2D human pose $b^p \in \mathbb{R}^{k\times2}$ to encode the HOI spatial information, where $k$ is the number of human keypoints. We normalize the coordinates of bounding boxes and pose into feature vector $f^{sp}$. In detail, denote the tight union bounding box of $b^h,b^o$ as $b^u\in \mathbb{R}^{2\times2}$ (represented by the upper-left and bottom-right points), we get $f^{sp}$=$\rm{Flatten}(\frac{\rm{concat}(b^h,b^o,b^p)-b^u[0,:]}{b^u[1,:]-b^u[0,:]})$, where $\rm{Flatten}(\cdot)$ indicates the transformation to a vector. 
Finally, $f^{sp}$ is fed into an MLP to classify the verbs. 

\noindent{\bf Uncertainty Quantification Module}
Moreover, we corporate an additional MLP as uncertainty quantification module in each stream, following the aleatoric uncertainty estimation in \cite{kendall2017uncertainties}. Each data point is assumed to possess a fixed uncertainty. Then, each stream not only outputs a logit $s$ for verb $v \in V$, but also estimates the uncertainty by log variance $e$. 
Thus, loss for a sample of verb $v$ is $L^s$=$||\frac{\sigma(s)-y}{\exp(e)}||^2 + \frac{1}{2}e$ following the logistic regression setting in \cite{kendall2017uncertainties}, where $y$ indicates whether sample has verb $v$. Concurrent with the correct prediction, the loss encourages the model to have higher uncertainty on unfamiliar samples while avoiding being uncertain to all the samples. Overall, the model would output higher uncertainty for samples of unfamiliar object categories, thus reduce the overconfident mistakes and advance generalization.

\noindent{\bf Calibration-Aware Unified Inference.}
With \textit{separately} trained three streams, we need to unify them into one final result without loss of immunity and detection performance. 
Given the inherent compositional characteristic of HOI, it is crucial to explore how to unify multi-stream knowledge. 
To this end, instead of simple addition or multiplication~\cite{interactiveness,gao2018ican}, we propose calibration-aware unified inference as shown in Fig.~\ref{Figure:overview}, which exploits the learned uncertainty quantification, imports more flexibility, and preserves unique advantages of each stream. Denote the output logit vectors of three streams as $s^*$ and the log variance vectors as $e^*$, where $*$ is $h, o$ or $sp$. 
First, separate calibration is imposed on three streams. Intuitively, the \textbf{more uncertain} a prediction is, the \textbf{less} it should contribute to the final result. 
Thus, the calibrated prediction of the streams are formulated as $p^*$=$Sigmoid(\frac{w^*\cdot s^* + c^*}{\exp(e^*)}) \cdot det^h \cdot det^o$, $w^*$, $c^*$ are learnable scaling parameters following \cite{platt1999probabilistic}, and $det^h$, $det^o$ are the object detection confidences from the object detector. 
The parameters are trained to minimize the BCE loss as Eq.~\ref{eq:cross-entropy} on the \textbf{validation} set, thus helping each separate stream to achieve well-calibration. The objective is formulated as $L^*=L_{ent}(p^*, y)$, $y\in [0,1]^{|V|}$ is the label vector.
Second, we impose a distributional agreement loss among the three streams, which is formulated as $L_{agree}=|\rm{mean}\it{(p^h_j - p^o_j)}| + |\rm{mean}\it{(p^o_j - p^{sp}_j)}| + |\rm{mean}\it{(p^{sp}_j - p^h_j)}|$, where $p^h_j$ denotes all the predictions of $p^h$ for verb $v_j$, the same is for $p^o_j$ and $p^{sp}_j$. This constraint does not require different streams output same value for the same data point like \cite{djrn}. Instead, it expects different streams to have similar average outcomes for a set of data points, which is more flexible.
Finally, to achieve better overall performance and calibration, we learn fusion factors $f^h$, $f^o$ and $f^{sp}$ on the validation set, where $f^h, f^o, f^{sp}>0$ and $f^h$ $+$ $f^o$ $+$ $f^{sp}$ = $1$. The final unified inference score is formulated as $\mathcal{P} $=$ f^h \cdot p^h + f^o \cdot p^o + f^{sp} \cdot p^{sp}$. We use a BCE loss $L_{uni}$=$-(\log(1 - \mathcal{P})(1 - y)+ \log(\mathcal{P})y)$ as the objective.
The total calibration loss is formulated as $L_{cal} $=$ \beta(L^h$+$L^o$+$L^{sp})$+$ \gamma L_{agree} $+$ L_{uni}$, where $\beta$ and $\gamma$ are both weighting factors. The parameters of each stream are frozen during calibration.

Overall, we introduce OC-immunity to the multi-stream structure, encouraging it to be insensitive to object categories while keeping representative to verbs. Thus, our model is robust against different object distributions. Meanwhile, the uncertainty quantification module provides clues on how much the model output could be trusted. It prevents the model from committing over-confident mistakes, boosting its generalization on unfamiliar objects. Finally, the proposed calibration-aware unified inference helps exploit most of each stream in the result, bringing better generalization.

\section{Experiment}
In this section, we first introduce our uncertainty-guided training strategy. 
Then, we introduce the adopted datasets, metrics and implementation details. 
Next, we compare our method with previous state-of-the-arts on multiple HOI datasets~\cite{hicodet,djrn,li2019hake}.
At last, ablation studies are conducted.

\subsection{Uncertainty-Guided Training}
\label{sec:ssl}
\begin{figure}[!t]
	\begin{center}
		\includegraphics[width=0.4\textwidth]{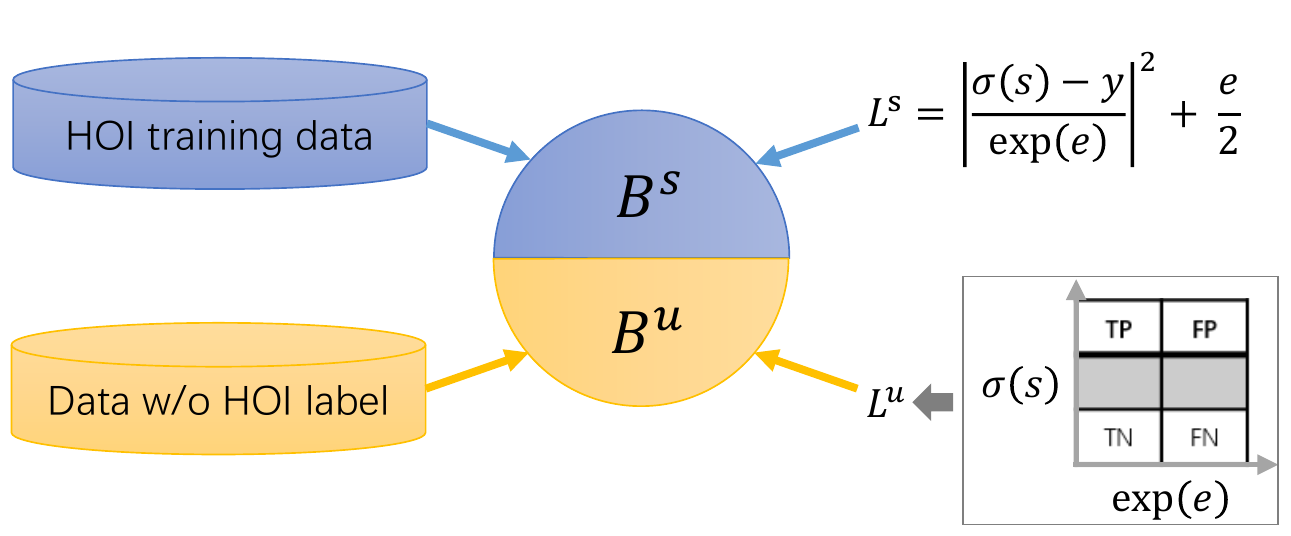}
	\end{center}
	\caption{With uncertainty estimation, we are enabled to utilize data without HOI label as an extra benefit.}
	\label{Figure:scheme}
	\vspace{-0.5cm}
\end{figure}
Besides conventional fully-supervised training, we could enhance our model with data \textbf{without HOI labels} as an extra benefit. Previous methods fail to exploit unlabeled data since only verb predictions are provided, but no clue is there for correctness. While our uncertainty quantification could additionally provide \textit{ambiguous but trustworthy} pseudo label of correctness, enabling the model to generate pseudo labels for the unlabeled data and further boost both verb classification and uncertainty quantification in a self-training manner.
We collect data from validation set of OpenImage~\cite{OpenImages} as extra unlabeled data and extract the required features for different streams as in Sec.~\ref{sec:method}. 
With these, we can operate uncertainty-guided training using unlabeled data as shown in Fig.~\ref{Figure:scheme}. 

\begin{figure}[!t]
	\begin{center}
		\includegraphics[width=0.45\textwidth]{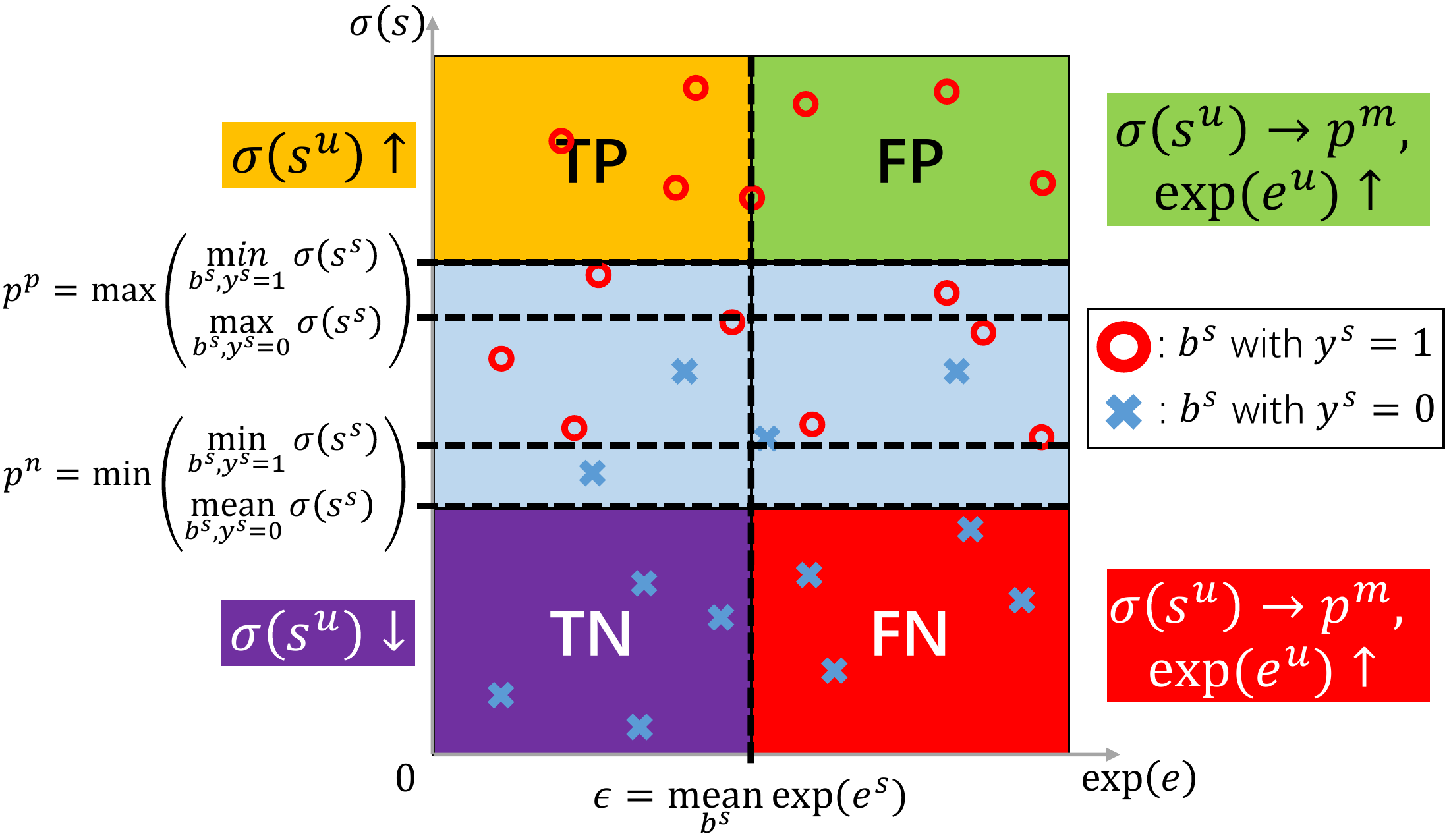}
	\end{center}
	\caption{We divide unlabeled samples into TP, TN, FP, and FN according to the output confidence and uncertainty. Based on this, loss is designed to tune the model.}
	\label{Figure:semi-sup}
	\vspace{-0.3cm}
\end{figure}

A mini-batch $B$ is \textit{mixed} of labeled data $B^s$=$\{b^s\}$ and unlabeled data $B^u$=$\{b^u\}$. 
For labeled sample $b^s \in B^s$ and verb $v$, the model outputs logit $s^s$ and log variance $e^s$. 
Thus, the loss for $b^s$ is calculated following $L^s$ as stated before.
For unlabeled sample $b^u \in B^u$ and verb $v$, we define the result evaluation (true positive, false positive, true negative and false negative) with respect to $B^s$, then calculate loss $L^u$ for them as shown in Fig.~\ref{Figure:semi-sup}. First, we compute thresholds $p^p$, $p^n$, $p^m$, $\epsilon$ as Eq.~\ref{eq:pt}-\ref{eq:var}, where $\sigma(x)=\frac{1}{1+\exp(-x)}$:
\begin{eqnarray}
   \label{eq:pt}
   p^p&=&\max(\mathop{\rm{mean}}_{b^s, y^s=1}(\sigma(s^s)), \mathop{\max}_{b^s, y^s=0}(\sigma(s^s))), \\
   \label{eq:pn}
   p^n&=&\min(\mathop{\rm{mean}}_{b^s, y^s=0}(\sigma(s^s)), \mathop{\min}_{b^s, y^s=1}(\sigma(s^s))), \\ 
   \label{eq:pmid}
   p^m&=&\frac{1}{2}(p^p + p^n), \\ 
   \label{eq:var}
   \epsilon&=&\mathop{\rm{mean}}_{b^s}(\exp(e^s)). 
\end{eqnarray}
Then, for unlabeled instance $b^u$, if $\sigma(s^u)$ is higher than $\sigma(s^s)$ of all labeled negative samples ($b^s\in B^s, y^s$=$0$) and the mean $\sigma(s^s)$ of all labeled positive samples ($b^s\in B^s, y^s$ = $1$), formulated as $\sigma(s^u)>p^p$, we define it as a positive sample. 
Similarly, if $\sigma(s^u)$ is lower than $\sigma(s^s)$ of all labeled positive samples and the mean $\sigma(s^s)$ of all labeled negative samples, we define it as a negative sample. 
Next, if the variance $\exp(e^u)$ is lower than the mean variance of all labeled samples $b^s\in B^s$, we define it as a true sample, otherwise a false sample. 
With these definitions, we construct loss based on BCE loss (Eq.~\ref{eq:cross-entropy}) for the unlabeled samples as Eq.~\ref{eq:loss-unlabel}:
\begin{equation}
\label{eq:cross-entropy}
L_{e}(p, y) = -(\log(1 - p)(1 - y)+ \log(p)y),
\end{equation}
\begin{equation}
\label{eq:loss-unlabel}
L^u=\left\{
\begin{array}{rcl}
L_{e}(\sigma(s^u), p^m) - e^u,& \sigma(s^u)>p^p, \sigma(e^u) > \epsilon,\\
L_{e}(\sigma(s^u), 1)        ,& \sigma(s^u)>p^p, \sigma(e^u) < \epsilon,\\
L_{e}(\sigma(s^u), p^m) - e^u,& \sigma(s^u)<p^n, \sigma(e^u) > \epsilon,\\
L_{e}(\sigma(s^u), 0)        ,& \sigma(s^u)<p^n, \sigma(e^u) < \epsilon,\\
0                            ,& otherwise
\end{array} \right.
\end{equation} 
The loss is straightforward: 
For TP/TN samples, we suppose these samples are labeled with corresponding $y^u$, and impose BCE loss on them, while we neither punish nor encourage the uncertainty estimation to avoid bias.
For FP/FN samples, we expect the model to be uncertain and the prediction to be mediocre. 
For the samples defined neither positive nor negative, we identify them as \textit{unfamiliar} samples and impose no loss on them, since they receive expected mediocre prediction.
The overall loss of mini-batch $B$ for each stream is represented as $L_{ugt}=\frac{1}{|V|}(\mathop{\sum}\limits_{b^s\in B^s} \mathop{\sum}\limits_{v\in V} \frac{L^s}{|B^s|} +  \mathop{\sum}\limits_{b^u\in B^u} \mathop{\sum}\limits_{v\in V} \frac{\alpha L^u}{|B^u|})$, where $\alpha$ is a weighting factor. 
With the designed loss, our model could generate trustworthy pseudo-label, regularize the verb classification and uncertainty estimation, and further boost generalization.
The corresponding analysis is detailed in Sec.~\ref{sec:result}-\ref{sec:analysis_pseudo}.

\subsection{Dataset and Metric}
\label{sec:dataset}
We adopt three large-scale HOI detection benchmarks: HICO-DET~\cite{hicodet}, Ambiguous-HOI~\cite{djrn}, and self-compiled HAKE~\cite{li2019hake} test set. The detailed statistics of the datasets are illustrated in the supplementary material.
For all three datasets, we evaluate our HICO-DET trained model using mAP following \citet{hicodet}: true positive should accurately locate human/object and classify the verb. The proposed mPD is evaluated under Default mode on verbs available for multiple object categories. 

\subsection{Implementation Details} 
\label{sec:implementation}
We re-split HICO-DET training set into train set and validation set, roughly a 6:4 split. And we collect 10,434 images from OpenImage validation set~\cite{OpenImages} as extra unlabeled data, containing 43,553 human instances, 77,485 object instances and 636,649 pairs. 
In the following, `train` refers to training on the train set and unlabeled data, while `calibrate` refers to calibrating on the validation set. 
All three streams are separately trained by SGD optimizer with learning rate of 7e-3. We train human stream for 50 epochs, the other two for 40 epochs. The unified calibration takes 2 epochs with SGD optimizer, learning rate of 1e-3, $\beta$ = $1$, and $\gamma$ = $0.1$.
All experiments are conducted on a single NVIDIA Titan Xp GPU.
Please refer to the supplementary material for more details.

\subsection{Results on Conventional HOI Detection}
\label{sec:result}
\begin{table}[t] 
    \centering
    \resizebox{\linewidth}{!}{\setlength{\tabcolsep}{0.8mm}{
    \begin{tabular}{l c c c c c c c}
        \hline
                                      & mPD $\downarrow$ & \multicolumn{3}{c}{mAP Default $\uparrow$}    &\multicolumn{3}{c}{mAP Known Object $\uparrow$} \\
        Method                        & & Full & Rare & Non-Rare & Full & Rare & Non-Rare \\
        \hline
        \hline
        \citet{Shen2018Scaling}       & -      & 6.46  & 4.24  & 7.12  & - & - & -             \\
        HO-RCNN~\cite{hicodet}        & -      & 7.81  & 5.37  & 8.54  & 10.41 & 8.94  & 10.85 \\
        \citet{Gkioxari2017Detecting} & -      & 9.94  & 7.16  & 10.77 & - & - & -             \\
        GPNN~\cite{gpnn}              & -      & 13.11 & 9.34  & 14.23 & - & - & -             \\
        \citet{knowledge}             & -      & 14.70 & 13.26 & 15.13 & - & - & -             \\ 
        iCAN~\cite{gao2018ican}       & 0.4699 & 14.84 & 10.45 & 16.15 & 16.26 & 11.33 & 17.73 \\
        \citet{wang2019deep}          & -      & 16.24 & 11.16 & 17.75 & 17.73 & 12.78 & 19.21 \\
        TIN~\cite{interactiveness}    & 0.4313 & 17.03 & 13.42 & 18.11 & 19.17 & 15.51 & 20.26 \\
        No-Frills~\cite{NoFrills}     & -      & 17.18 & 12.17 & 18.68 & - & - & - \\
        \citet{zhou2019relation}      & -      & 17.35 & 12.78 & 18.71 & - & - & - \\
        PMFNet~\cite{pmfnet}          & -      & 17.46 & 15.65 & 18.00 & 20.34 & 17.47 & 21.20 \\
        \citet{analogy}               & 0.4314 & 19.40 & 14.60 & 20.90 & -     & -     & -     \\
        \citet{vsgnet}                & - & 19.80 & 16.05 & 20.91 & - & - & - \\
        DJ-RN~\cite{djrn}             & \underline{0.4121} & \underline{21.34} & \underline{18.53} & \underline{22.18} & \underline{23.69} & \underline{20.64} & \underline{24.60} \\ 
        \textbf{Ours}                 & \textbf{0.3836} & \textbf{21.95} & \textbf{20.89} & \textbf{22.27} & \textbf{24.79} & \textbf{23.82} & \textbf{25.08}\\ 
        \hline
        PPDM~\cite{ppdm}           & 0.3930 & 21.73 & 13.78 & 24.10 & 24.58 & 16.65 & 26.84 \\
        \citet{functional}         & -      & 21.96 & 16.43 & 23.62 & - & - & - \\
        VCL~\cite{vcl}             & 0.4106 & 23.63 & 17.21 & 25.55 & 25.98 & 19.12 & 28.03 \\
        IDN~\cite{li2020hoi}       & \textbf{0.3876} & \textbf{26.29} & \underline{22.61} & \textbf{27.39} & \textbf{28.24} & \textbf{24.47} & \textbf{29.37} \\
        \textbf{Ours}              & \underline{0.3905} & \underline{25.44} & \textbf{23.03} & \underline{26.16} & \underline{27.24} & \underline{24.32} & \underline{28.11} \\ 
        \hline
        iCAN~\cite{gao2018ican}    & 0.3441 & 33.38 & 21.43 & 36.95 & - & - & - \\
        TIN~\cite{interactiveness} & 0.3421 & 34.26 & 22.90 & 37.65 & - & - & - \\ 
        \citet{analogy}            & 0.3203 & 34.35 & 27.57 & 36.38 & - & - & - \\ 
        VCL~\cite{vcl}             & \underline{0.3097} & 38.97 & 29.99 & 41.65 & - & - & - \\ 
        IDN~\cite{li2020hoi}       & - & \textbf{43.98} & \textbf{40.27} & \textbf{45.09} & - & - & - \\
        \textbf{Ours}              & \textbf{0.2971} & \underline{41.32} & \underline{35.57} & \underline{43.03} & - & - & - \\ 
        \hline
      \end{tabular}}}
        \caption{Results on HICO-DET~\cite{hicodet}. The first part adopted COCO pre-trained detector. HICO-DET fine-tuned detector is used in the second part. GT human-object pair boxes are used in the last part.}
        \label{tab:hicodet} 
\vspace{-0.3cm}
\end{table}
\noindent{\bf HICO-DET}: Quantitative results are demonstrated in Tab.~\ref{tab:hicodet}, compared with previous state-of-the-art methods using different object detectors. 
The results are evaluated following \citet{hicodet}: Full (600 HOIs), Rare (138 HOIs), and Non-Rare (462 HOIs) in Default and Known Object mode. 
Also, we compare mPD with some open-sourced algorithms in Tab.~\ref{tab:hicodet}. As shown, our method provides an impressive mAP of 21.95 (Default Full) and outperforms all previous algorithms on mPD (0.3836) with COCO~\cite{coco} pre-trained object detector. And it achieves similar performance with IDN~\cite{li2020hoi} using HICO-DET fine-tuned object detector and GT human-object pairs. It is even comparable with very recent transformer-based methods (e.g., \citet{kim2021hotr}: mAP 25.73, mPD 0.3978) with much larger capacity.

\noindent{\bf Ambiguous-HOI}~\cite{djrn} is adopted to further evaluate our model on unfamiliar data.
As shown in Tab.~\ref{tab:ambi}, our model provides competitive mAP and considerable improvement on mPD over previous state-of-the-arts, proving its robustness against object category distribution.

\noindent{\bf HAKE test set}: Results are illustrated in Tab.~\ref{tab:ambi}. Some previous open-sourced SOTA are compared. We could observe the superior performance of our method even with domain shift, demonstrating the generalization ability of our model.
\begin{table}[!t]
\centering
\resizebox{0.45\textwidth}{!}{
\begin{tabular}{l c c c c}
\hline
                           & \multicolumn{2}{c}{Ambiguous-HOI}    &\multicolumn{2}{c}{HAKE test set} \\
Method                     & mPD $\downarrow$ & mAP $\uparrow$   & mPD $\downarrow$   & mAP $\uparrow$\\
\hline
\hline
iCAN~\cite{gao2018ican}    & 0.5060 & 8.14  & 0.5324 & 9.66  \\ 
TIN~\cite{interactiveness} & 0.4890 & 8.22  & 0.4792 & 10.45 \\
\citet{analogy}            & 0.4935 & 9.72  & 0.5132 & 12.03 \\
DJ-RN~\cite{djrn}          & 0.4800 & 10.37 & - & - \\
\hline
Ours                       & \textbf{0.4715} & \textbf{10.45} & \textbf{0.4736} & \textbf{14.26}\\
\hline
\end{tabular}}
\caption{Results comparison on Ambiguous-HOI~\cite{djrn} and newly designed test set from HAKE.} 
\vspace{-0.3cm}
\label{tab:ambi}
\end{table}
\vspace{-0.3cm}

\subsection{Results on Zero-shot HOI Detection}
To demonstrate the generalization ability of our method, we evaluate the performance of our method under zero-shot settings. 120 of the 600 HOI categories in HICO-DET~\cite{hico} are selected as unseen categories~\cite{Shen2018Scaling,vcl}. We adopt the non-rare first selection~\cite{vcl}, which is more difficult. Instances of the unseen categories are removed during training, while the test set is unchanged. mAP is reported under three settings (Full, Seen, Unseen) with Default mode. mPD is evaluated for open-sourced methods~\cite{vcl} and our method. Comparison is conducted among several previous methods under the same setting.
As shown in Tab.~\ref{tab:zero}, our method considerably outperforms previous methods.
More experiments are in the supplementary material.

\begin{table}[t]
    \centering
    \resizebox{0.4\textwidth}{!}{
    \begin{tabular}{l c c c c}
        \hline
                                & mPD $\downarrow$ & \multicolumn{3}{c}{mAP $\uparrow$}\\
        Method                   &      & Full  & Seen & Unseen\\
        \hline
        \hline
        \citet{Shen2018Scaling} & -      & 6.26  & -     & 5.62\\
        VCL~\cite{vcl}          & 0.4877 & 12.76 & 13.67 & 9.13\\
        \textbf{Ours}           & \textbf{0.4415} & \textbf{14.80} & \textbf{15.70} & \textbf{11.21} \\ 
        \hline
        \citet{functional}      & -       & 12.26 & 12.60 & 10.93\\
        VCL~\cite{vcl}          & 0.4531  & 18.06 & 18.52 & 16.22\\
        \textbf{Ours}           & \textbf{0.4194} & \textbf{19.85} & \textbf{20.23} & \textbf{18.34} \\ 
        \hline
      \end{tabular}}
      \vspace{-0.2cm}
        \caption{Results of zero-shot HOI detection. The first part adopted COCO pre-trained detector. HICO-DET fine-tuned detector is used for the second part.}
        \label{tab:zero} 
\vspace{-0.4cm}
\end{table}
\subsection{Visualization}
\label{sec:vis}
\begin{figure}[!t]
	\begin{center}
		\includegraphics[width=0.4\textwidth]{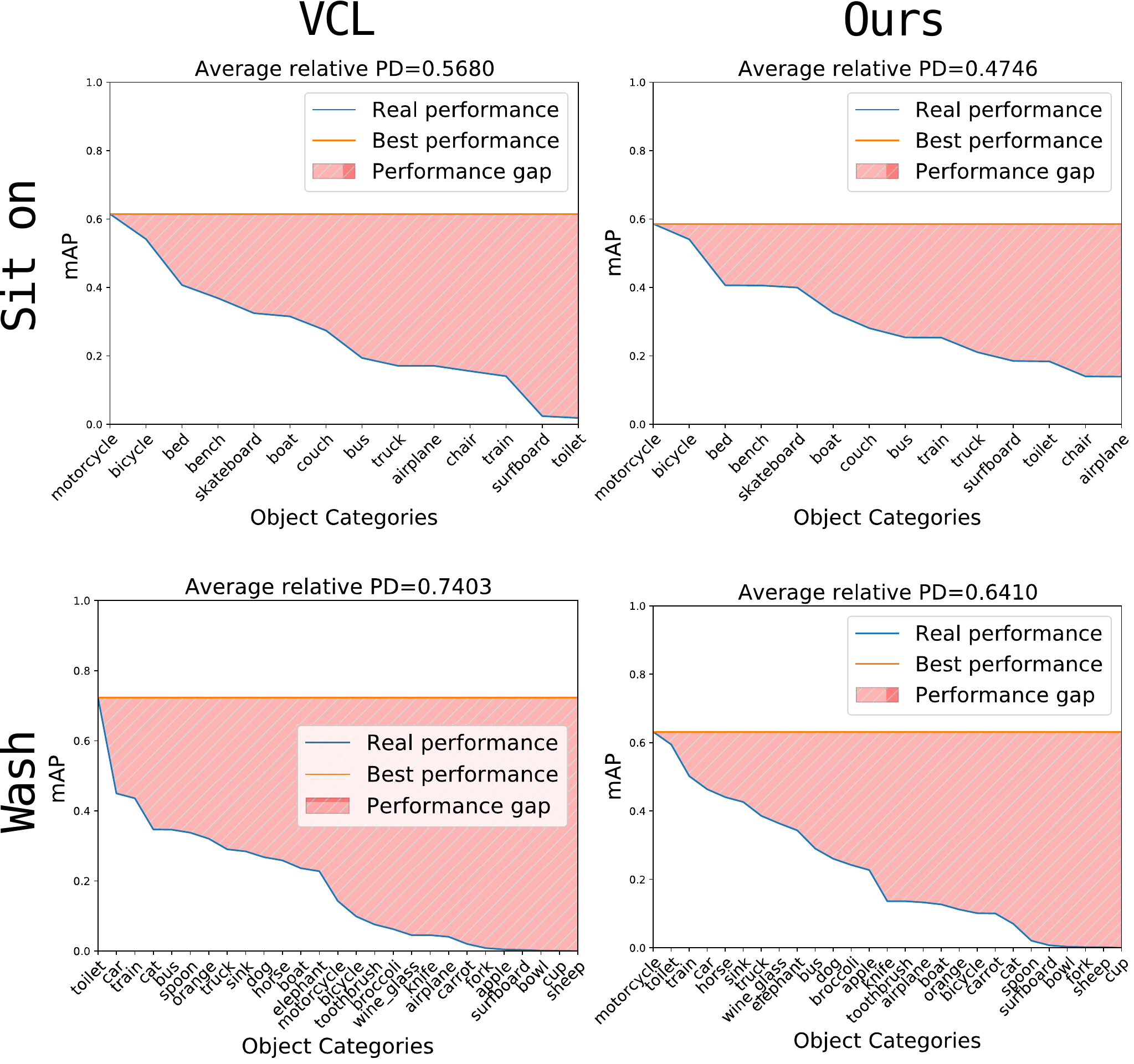}
	\end{center}
\vspace{-0.4cm}
	\caption{mPD comparison of VCL and ours.}
	\label{Figure:mpd}
\vspace{-0.4cm}
\end{figure}
We visualize mPD comparison of VCL~\cite{vcl} and ours in Fig.~\ref{Figure:mpd}. As shown, our method shows similar or even lower best AP while still outperforming VCL~\cite{vcl} by providing less performance degradation, implying that we manage to achieve a better trade-off. More visualizations are included in the supplementary material.

\subsection{Ablation Study}
\label{sec:ablation}
We conduct ablation studies on HICO-DET~\cite{hicodet} with COCO~\cite{coco} pre-trained Faster-RCNN~\cite{faster}. For more ablation studies please refer to the supplementary material.

\noindent{\bf OC-immune learning (OIL):}. We remove the synthesizer and train the verb classifier directly with raw data. As shown, the performance for Non-rare set barely hurts, while both mPD and Rare mAP decrease substantially. This sustains the effectiveness of OIL in mitigating the performance gap between different objects with same verb.

\noindent{\bf Uncertainty quantification module (UQM):} Its removal results in a 0.52 mAP drop and marginal mPD increase. The performance on Rare hurts more than that on Non-Rare, indicating the importance of UQM for unfamiliar samples.

\noindent{\bf Calibration-aware unified inference (CUI):} If we jump the calibration stage, and directly fuse the three streams, mAP suffers by 0.79, proving the crucial role of CUI.

\noindent{\bf Extra data:} Without the extra data, we get mAP of 20.05 and mPD of 0.3987, which is still competitive. In additional, we achieve 25.44 mAP and 0.3905 mPD using HICO-DET fine-tuned detector with no extra data, which is comparable with even very recent transformer methods (\citet{kim2021hotr} with mAP of 25.73 and mPD of 0.3978).

\noindent{\bf Different streams:} Spatial stream provides the best individual result, despite its simplicity. 
The three streams on their own do not provide superior performance, while the combination stands out. Meanwhile, individual streams are of high mPD, while the final result gives significantly lower mPD, indicating the importance of combination.
Impressively, object stream performs better on Rare set than on Non-Rare set, implying the effectiveness of the OC-immune learning.

\noindent{\bf $L_{agree}$:} We evaluate $L_{agree}$ by removing it and replace it with that in \cite{djrn}. As shown, $L_{agree}$ mostly benefits performance on Rare set, and our distributional agreement loss is slightly better than the point-level agreement.

\begin{table}[!t]
    \centering
      \resizebox{0.8\linewidth}{!}{
        \begin{tabular}{l  c  c  c  c}
          \hline
                         & mPD $\downarrow$ & \multicolumn{3}{c}{mAP Default $\uparrow$}   \\
          Method         & & Full & Rare & Non-Rare   \\
          \hline
          \hline
          Ours           & \textbf{0.3836} & \textbf{21.95} & \textbf{20.89} & \textbf{22.27} \\ 
          \hline
          w/o OIL        & 0.4001 & 21.34 & 18.46 & 22.20 \\
          w/o UQM        & 0.3860 & 21.43 & 19.86 & 21.90 \\
          w/o CUI        & 0.3871 & 21.16 & 20.31 & 21.42 \\
          w/o Extra Data & 0.3987 & 20.05 & 18.66 & 20.47 \\
          \hline
          Human stream   & 0.4927 & 10.81 & 10.09 & 11.03 \\  
          Object stream  & 0.4891 & 10.29 & 11.15 & 10.03 \\  
          Spatial stream & 0.4487 & 15.61 & 13.81 & 16.15 \\  
          \hline
          w/o $L_{agree}$               & 0.3860 & 21.71 & 20.20 & 22.17 \\
          $L_{agree}$ from \citet{djrn} & 0.3855 & 21.81 & 20.50 & 22.21 \\  
          \hline
        \end{tabular}}
\vspace{-0.3cm}
      \caption{Ablation study results.}
      \label{tab:ablation}
      \vspace{-0.3cm}
\end{table}

\subsection{Analysis on Uncertainty-guided Training}
\label{sec:analysis_pseudo}
To verify the definitions statistically, we select 400 samples from HICO-DET~\cite{hicodet} training set as labeled data and 400 samples from HICO-DET test set as unlabeled data, then we generate pseudo labels for the unlabeled data and calculate their quality as shown in Tab.~\ref{tab:def}. Most GT TP and FP samples, which are of main consideration in HOI, correspond well with the pseudo labels. 

\begin{table}[!t]
    \centering
      \resizebox{0.6\linewidth}{!}{
    \begin{tabular}{c|c|c}
        \hline
         & Pseudo TP & Pseudo FP \\
        \hline
        Percent of GT TP & \textbf{76.98}\% & 48.48\% \\
        Percent of GT FP & 23.02\% & \textbf{51.52}\% \\
        \hline
    \end{tabular}}
\vspace{-0.3cm}
    \caption{Pseudo label distribution.}
    \label{tab:def}
    \vspace{-0.4cm}
\end{table}

\section{Conclusion}
In this paper, we proposed a novel metric mPD as a complement of mAP for measurement of generalization. Based on mPD, we raised to seek generalization via OC-immunity, and designed a new OC-immune network, achieving impressive improvements for both conventional and zero-shot generalization HOI detection.

\section*{Acknowledgments}
This work is supported in part by the National Key R\&D Program of China, No. 2017YFA0700800, Shanghai Municipal Science and Technology Major Project (2021SHZDZX0102), National Natural Science Foundation of China under Grants 61772332 and Shanghai Qi Zhi Institute, SHEITC (2018-RGZN-02046).

\appendix

\bibliography{aaai22}

\end{document}